\documentclass[a4paper, 10pt]{article}      

\usepackage[T1]{fontenc}
\usepackage[utf8]{inputenc}
\usepackage[pdftex]{graphicx}
\usepackage[margin=1in]{geometry}
\usepackage{authblk}

\usepackage{hyperref}
\title{\LARGE \bf
The STRANDS Project: Long-Term Autonomy in Everyday Environments
}

\title{The STRANDS Project: Long-Term Autonomy in Everyday Environments}
\author[1]{Nick Hawes}
\author[1]{Chris Burbridge}
\author[1]{Ferdian Jovan}
\author[1]{Lars Kunze}
\author[1]{Bruno Lacerda}
\author[1]{Lenka Mudrov\'a}
\author[1]{Jay Young}
\author[1]{Jeremy Wyatt}
\author[2]{Denise Hebesberger}
\author[2]{Tobias K\"ortner}
\author[3]{Rares Ambrus}
\author[3]{Nils Bore}
\author[3]{John Folkesson}
\author[3]{Patric Jensfelt} 
\author[4]{Lucas Beyer}
\author[4]{Alexander Hermans}
\author[4]{Bastian Leibe} 
\author[5]{Aitor Aldoma}
\author[5]{Thomas F{\"a}ulhammer} 
\author[5]{Michael Zillich} 
\author[5]{Markus Vincze} 
\author[6]{Eris Chinellato}

\author[7]{Muhannad Al-Omari}
\author[7]{Paul Duckworth}
\author[7]{Yiannis Gatsoulis}
\author[7]{David C. Hogg}
\author[7]{Anthony G. Cohn}
\author[8]{Christian Dondrup}
\author[8]{Jaime Pulido Fentanes}
\author[8]{Tom{\'a}{\v s} Krajn{\'i}k}
\author[8]{Jo\~ao M. Santos}
\author[8]{Tom Duckett}
\author[8]{Marc Hanheide} 

\affil[1]{Intelligent Robotics Lab, School of Computer Science, University of Birmingham, UK}%
\affil[2]{Akademie Fur Altersforschung Am Haus Der Barmherzigkeit, Austria; and Donau-Universitaet Krems, Austria}%
\affil[3]{Centre for Autonomous Systems, KTH Royal Institute of Technology, SE-100 44 Stockholm, Sweden}%
\affil[4]{Rheinisch-Westf\"alische Technische Hochschule Aachen, Germany}%
\affil[5]{Technische Universit\"at Wien, Austria}%
\affil[6]{Faculty of Science and Technology, Middlesex University London, UK}%
\affil[7]{University of Leeds, UK}%
\affil[8]{LCAS, University of Lincoln, UK}%

\begin{document}

\maketitle
\thispagestyle{empty}
\pagestyle{empty}



\section{Introduction}

Thanks to the efforts of the robotics and autonomous systems community, robots are becoming ever more capable. There is also an increasing demand from end-users for autonomous service robots that can operate in real environments for extended periods. In the STRANDS project\footnote{Spatio-Temporal Representations and Activities for Cognitive Control in Long-Term Scenarios, \url{http://strands-project.eu}.} we are tackling this demand head-on by integrating state-of-the-art artificial intelligence and robotics research into mobile service robots, and deploying these systems for long-term installations in security and care environments.
Over four deployments, our robots have been operational for a combined duration of 104 days autonomously performing end-user defined tasks, covering 116km in the process. In this article we describe the approach we have used to enable long-term autonomous operation in everyday environments, and how our robots are able to use their long run times to improve their own performance.

\section{Long-Term Autonomy in STRANDS}\label{sec:lta}

Autonomous robots come in a range of forms, for a range of applications. Across this range, long-term autonomy (LTA) has a variety of meanings. For example, NASA's Opportunity rover has been autonomous for over 10 years on the surface of Mars; wave gliders can autonomously monitor stretches of ocean for months at a time; and autonomous cars have completed journeys of thousands of kilometres. In this article we restrict our contributions to \emph{mobile robots operating in everyday, indoor environments} (e.g. offices, hospitals), capable of performing \emph{a variety of service tasks}. Across all the aforementioned robots there are commonalities in low-level, short-term control algorithms (e.g. closed-loop motor control). Beyond this, the algorithms used to provide long-term, task-specific autonomous capabilities, and the hardware these algorithms control, varies greatly, according to application and environmental requirements. The challenges that distinguish indoor service robots from the aforementioned examples relate to both their environment and their task capabilities. Indoor task environments are less physically risky than outdoor environments, but have a comparatively higher degree of short- to medium-term physical variability, e.g. people, doors and furniture moving (roads are similar, but traffic movement is generally more predictable and less frequently occluded). In terms of application requirements, multi-purpose service robots must be capable of predictable scheduled behaviour whilst also being retaskable on-demand with high availability, and must be able to navigate in relatively confined, dynamics environments. This is in contrast to the largely restricted-purpose systems mentioned above. Taken together the set of requirements for indoor service robots presents unique challenges, and thus LTA in this context warrants dedicated research.

Given the state of the art, we consider ``long-term'' for a mobile service robot to be at least multiple weeks of \emph{continuous} operation. In very general terms, such LTA operation requires that a robot's hardware and software is robust enough to failure to enable such operation. Such robustness can be provided by both design-time and run-time approaches. It is essential that LTA systems are able to actively manage consumable resources (e.g. battery) and that any autonomy-supporting capabilities (e.g. localisation) are not adversely affected by long run times. Whilst this latter point is common sense, and common practice in many other technologies (from operating systems to cars), it has only recently been considered in autonomous robotics.

One reason it is challenging to design a service robot to meet the requirements of LTA is the impossibility of anticipating all the situations in which it may find itself. However, if we can enable  robots to run for long periods, then they will have opportunities to learn about the structure and dynamics of such situations. By exploiting the results of such learning, the robots should be able to increase their robustness further, leading to a virtuous cycle of improved performance and greater autonomy. It is this latter point which motivates STRANDS: to go beyond robots which simply survive, to those that can improve their performance in the long term. It is in this context that this article makes its main contribution: a robotic software architecture (the STRANDS Core System) which was designed for LTA service robot applications, and evaluated across four end-user deployments. It contains a mix of common sense and novel elements which have enabled it to support over 100 days of autonomous operation. This article is the first time all of these elements have been presented together, and contains the first presentation of metrics describing performance across deployments. Our approach is inspired by the work of Willow Garage~\cite{Marder-Eppstein:2010} and the CoBot project~\cite{1000kchallenge}, plus the pioneering work on systems like Rhino and Minerva (e.g.~\cite{Minerva}). 
What distinguishes our work from these is the combination of multiple service capabilities, in a single system capable of weeks or more of continuous autonomous operation, in dynamic indoor environments, whilst using various forms of learning to improve system performance. Many other projects address one or two of these elements, but not all four simultaneously.


\section{Application Scenarios}

\begin{figure*}
\begin{center}
\includegraphics[width=\columnwidth]{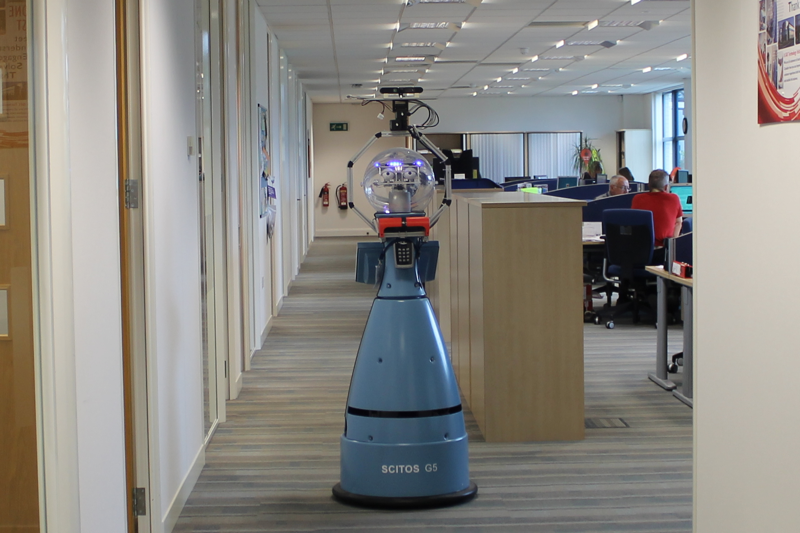}
\includegraphics[width=\columnwidth]{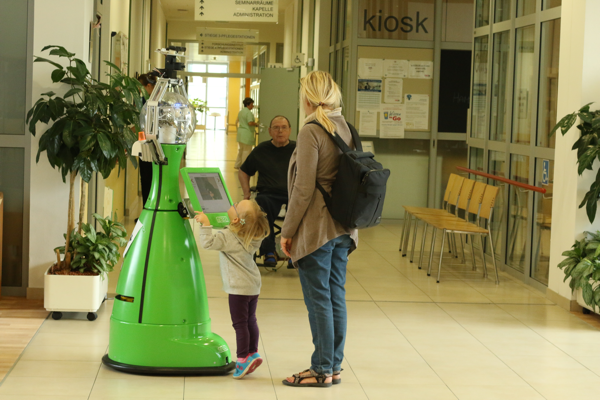}
\end{center}
\caption{Two of the STRANDS MetraLabs SCITOS A5s in their application environments. On the left is the robot \emph{Bob} at G4S's Challenge House in Tewkesbury, UK. On the right is the robot \emph{Henry} in the reception of Haus der Barmherzigkeit, Vienna. \label{fig:robots}}
\end{figure*}

To ensure our research is able meet the demands of end users, our work is evaluated in two application scenarios: security and care. Space does not permit a detailed explanation of the tasks in each scenario. Instead we include citations to further information on the tasks and technology from each scenario.
 
Our security scenario is developed with G4S Technology.
The aim of this scenario is to have a robot monitoring an indoor office environment, generating alerts when it observes prohibited or unusual events. To date we have completed two security deployments in which a mobile robot routinely created models of the environment's 3D structure~\cite{ambrus_iros15_st_models}, objects~\cite{faeulhammer_object_learning_ral} and people~\cite{Duckworth:2016}; modelling their changes over time; and using these models to detect anomalous situations and patterns. For example, we have developed robot behaviours to: detect when a human moves through the environment in an unusual manner~\cite{Duckworth:2016}; build models of the arrangement of objects on desks~\cite{kunze14topdown}; and check whether fire exits have been left open. Long-term deployments are essential for these services in order to gather sufficient data to build appropriate models. 

Our care scenario is developed with the Akademie f\"ur Alterforschung at the Haus der Barmherzigkeit (HdB).
In this scenario, the robot supports staff and patients in a large elderly care facility.
To date we have completed two care deployments in which a mobile robot: guides visitors; provides information to residents; and assists in walking-based therapies. In the care scenario the robot serves users more directly, and therefore long-term system robustness is crucial, as is adapting to the routines of the facility. For more information on this scenario see~\cite{Hebesberger2016hri,Hebesberger2015}.

\section{Robot Technology}

The systems reported in this paper are developed in ROS, available under open source licenses, and binary packaged for Ubuntu LTS.\footnote{See \url{http://strands-project.eu/software.html}.}
Whilst the majority of our work is platform neutral, all our deployed systems are based on the MetraLabs SCITOS A5 (see Figure~\ref{fig:robots}).
This is an industry-standard mobile robot capable of the long run times (12 hours on one charge) and autonomous charging. Our robots each have SICK S300 lasers in their bases (for localisation, leg detection etc.), and two Asus Xtion PRO RGB-D cameras: one at chest height pointing downward (for obstacle avoidance), the other on a pan-tilt unit (PTU) above the robot's head. The SCITOS has an embedded Intel Core i7 PC with 8GB RAM to which we have networked two additional PCs each with an i7 and 16 GB RAM.

\section{The Core STRANDS System}\label{sec:core}


\begin{figure}
\begin{center}
\includegraphics[width=\columnwidth]{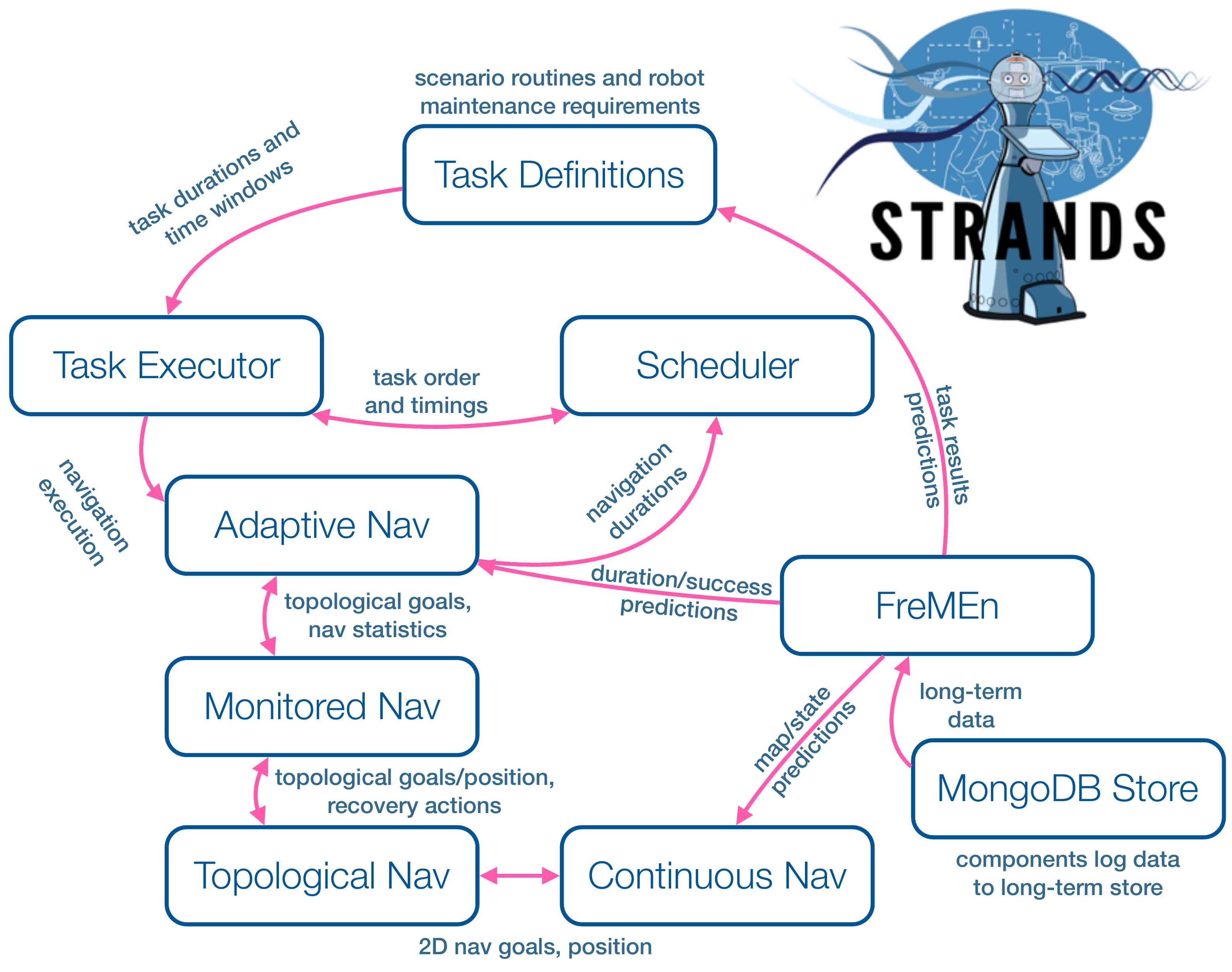}
\end{center}
\caption{Schematic overview of the Core STRANDS System.\label{fig:system_overview}}
\end{figure}

The \emph{STRANDS Core System} (Figure~\ref{fig:system_overview}) is an application-neutral architecture for LTA in mobile robots. It is a mix of widely-used components, plus components designed specifically for LTA. As mentioned above, hardware and software robustness is essential for LTA. Hardware robustness is beyond the scope of our research, thus we assume our software is running on an appropriate robot and computational platform. We address software component robustness through a mix of strategies. During development we encourage components to be designed in a way that makes the minimum assumptions about the existence of other components and services (e.g. by checking service existence before running). We also pay particular to error handling to ensure component-local errors and exceptions do not propagate unnecessarily. This allows components, and whole subsystems, to be brought up and down automatically. At run-time we use built-in ROS functionality to automatically relaunch crashed components, and try to run most subsystems only when required (saving CPU and power, and reducing opportunities for errors). We also use run-time topic monitoring to detect problems (e.g. low publish rates) and trigger component restarts. Finally, we run a continuous integration server that tests components and the whole system in isolation, on recorded data, and in simulation. The rest of this section summarises the STRANDS Core System, and provides references to additional technical details. 



The overall performance of a mobile robot is constrained by its localisation and navigation systems, so we use widely-adopted ROS packages to provide state-of-the-art performance.
 When deploying we build a fixed map from laser, localise in it using adaptive Monte Carlo localisation, and navigate using the dynamic window approach (DWA) over 3D obstacle information.\footnote{See \url{http://wiki.ros.org/navigation} for details on these techniques.} Whilst our use of a fixed map appears at odds with LTA in a dynamic environment, our environments are dominated by static features (e.g. walls), which prevent the robot's localisation performance from degrading. We also take advantage of the the robot regularly docking with a charging station by resetting the robot's position to this known location whilst docked. This limits localisation drift to that which can occur during time away from the dock. 


We manually build a topological map on top of the fixed continuous map.
We place topological nodes at key places in the environment for navigation (e.g. either side of a door) or for tasks (e.g. by a desk to observe).
The topological map from our 2015 security deployment is in Figure~\ref{fig:recovery_plot}.
Edges in the topological map are parametrised by the action required to move along them.
In addition to DWA navigation, our system can perform door passing, docking on to a charging station, and adaptive navigation near humans~\cite{Dondrup:2015}.

In our experience, navigation performance is major determiner of the autonomous run time of a mobile robot. This is because navigation failures (e.g. getting stuck near obstacles) can result in the robot being unable to return to its charging station. Thus the aforementioned elements of the STRANDS Core System support LTA in the following ways. First, by constraining the robot's movements to the topological map we are able to restrict navigation to known good areas of the environment. We additionally restrict movement by marking areas of the static map as `no go' zones which cannot be planned through. Despite these restrictions, navigation failures still occur due to environmental dynamics (e.g. people walking in front of the robot). Therefore edge traversals in the topological map are executed by a \emph{monitored navigation} layer that can perform a range of recovery actions in the event of failure (see Section~\ref{sec:monitored_nav}). Also, topological route planning and execution is one place where our core system adapts to long-term experience, as described in Section~\ref{sec:adaptive_nav}.



\begin{figure*}
\begin{center}
\includegraphics[width=\textwidth]{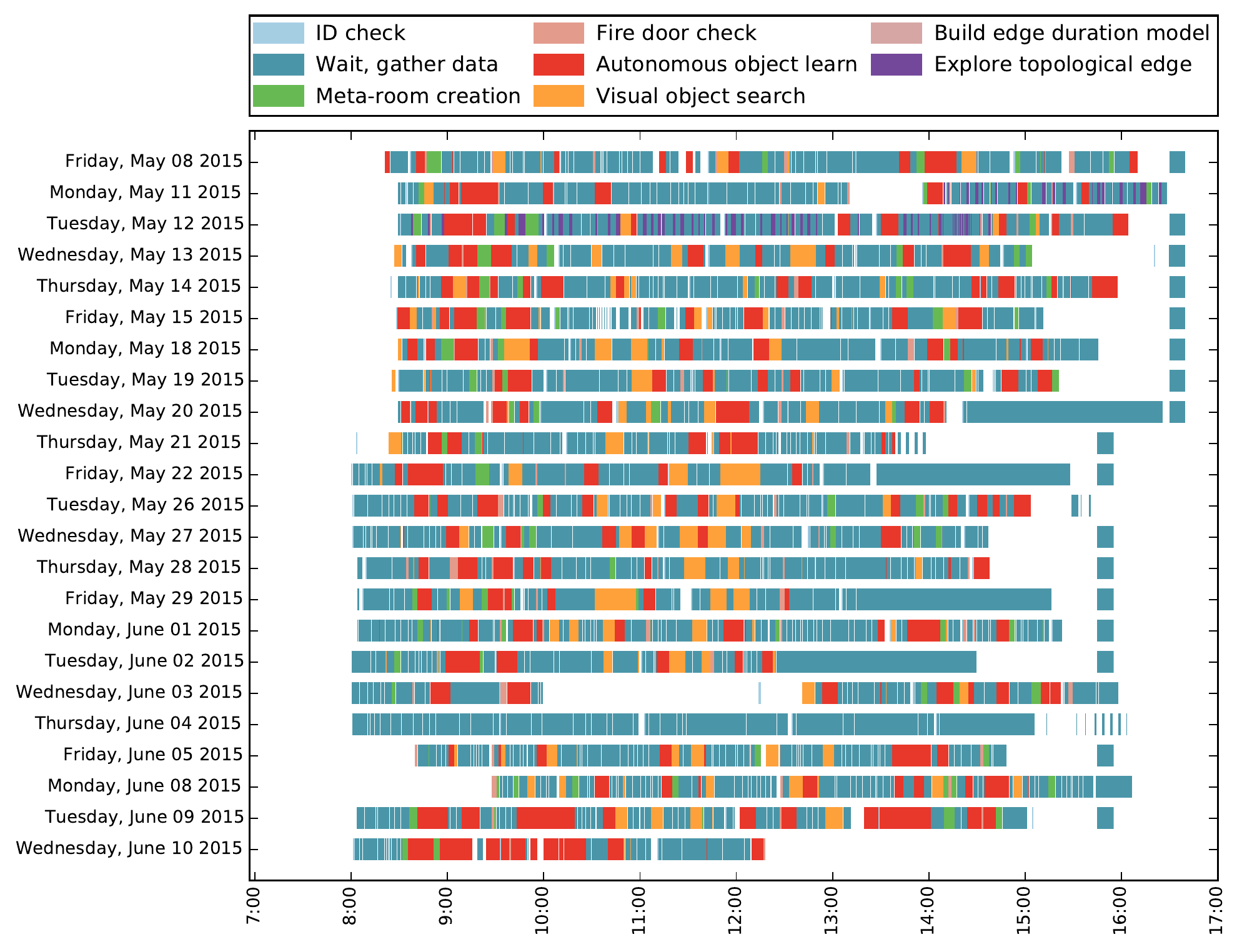}
\end{center}
\caption{A plot of the tasks performed by the robot during the 2015 security deployment. White space indicates that the robot is not performing any tasks. This indicates that the robot is charging or a failure has occurred.\label{fig:g4s_routine}}
\end{figure*}

The main unit of behaviour in our system is a \emph{task}.
Tasks represent something the robot can do (e.g. check whether a fire door is open, serve information via a GUI), and have an associated topological location, a maximum duration, and a time window for execution.
Our executive framework~\cite{Mudrova:2015} schedules tasks to be executed within their time windows, and manages task-directed navigation then execution.
To prevent task failures from interfering with long-term operation, our framework detects task time-outs and failures, then stops or restarts robot behaviours as necessary.  
Maintenance actions such as charging, batch learning and database backups are all handled as tasks, allowing the executive framework control of most of the robot's behaviour. This is essential for LTA as it enables the system to actively manage its limited resources.
A plot of tasks from the 2015 security deployment can be seen in Figure~\ref{fig:g4s_routine}.


Our system relies on separate pipelines for perceiving different elements of its environment: real-time multi-person RGB-D detection and tracking~\cite{hosseini14ICRA}; visual object instance and category modelling and recognition~\cite{Prankl:2015}; and 3D spatio-temporal mapping~\cite{ambrus_iros15_st_models}. This article does not cover our work on perceptually challenging tasks. Instead we refer readers to other papers where we have exploited these perception pipelines, e.g.~\cite{Dondrup:2015,faeulhammer_object_learning_ral,kunze14topdown}.


The data observed and generated (e.g. as inter-component communication) by an LTA system is crucial for both learning, and for monitoring and debugging the system. We therefore use tools based on MongoDB\footnote{\url{http://wiki.ros.org/mongodb_store}} to save ROS messages to a document-oriented database. Database contents (e.g. observations of doors being opened or closed) can then be interpreted by the Frequency Map Enhancement (FreMEn) component~\cite{fremen_2014_icra}, which integrates sparse and irregular observations into spatio-temporal models representing (pseudo-)periodic environment variations. These can be used to predict future environment states (see Section~\ref{sec:adaptive_nav}).



\section{Metrics}

So far we have performed two evaluation deployments for each of the security and care scenarios.
For each deployment we monitored overall system performance against two metrics: \emph{total system lifetime} (TSL), and \emph{autonomy percentage} (A\%).
TSL measures how long the system is available for autonomous operation, and is reset if the system experiences an unrecoverable failure, or needs an unrequested expert intervention (i.e. something which cannot easily be done by an end-user on site).
A\% measures the duration the system was actively performing tasks as a proportion of the time it was allowed to operate autonomously (which in our case is typically restricted to office hours).
The motivation of A\% is that it is trivial to achieve a long TSL if the system does nothing.
However, neither TSL nor A\% measure the quality of the services being provided. As this article focuses on LTA we restrict our presentation to the aforementioned, task-neutral but LTA-specific metrics. End-user evaluations of our systems' task-specific performance are ongoing, and will be published in the future (see ~\cite{Hebesberger2015,Hebesberger2016hri} for early evaluations from the care scenario).

\begin{table*}[ht]
\centering
\begin{tabular}{l||l|l|l|l||l}
				& Care 2014   & Security 2014  &  Care  2015 &  Security 2015 & Total\\
\hline
Total Distance Travelled		&  27.94km &	20.64km &	23.41km &	44.25km  & 116.24km \\
Total Tasks Completed			&  1985	&	963		&	865		&	4631	 & 8444 \\
Max TSL 			&  7d 3h &	6d 19h &	15d 6h &	28d 0h  & \\
Cumulative TSL 			& 20d 19h &	21d 0h &	27d 8h &	35d 3h  & 104d 7h \\
Individual Continuous Runs  			& 18 		&	18 		&	5 		&	2 & 43 \\
Autonomy Percentage (A\%) 			& 38.80\% 		&	18.27\% 		&	53.51\% 		&	51.10\% & \\
\end{tabular}
\caption{LTA metrics from the first four STRANDS system deployments.}
\label{tab:metrics}
\end{table*}

Table~\ref{tab:metrics} presents our systems' LTA performance so far.
In 2014 we aimed for 15 days TSL.
However the longest run we achieved was seven days. 
Most of our system failures were caused by the lack of robustness of our initial software, leading to unrecoverable component behaviour (crashes or deadlock states).
This was fixed for our 2015 deployments by following the development approaches outlined in Section~\ref{sec:core}.
%
In 2015 we targeted 30 days TSL, coming close with 28 days in the security deployment.
This long run was terminated by the robot's motors not responding to commands, an issue which has since been fixed by a firmware update. In the 2015 deployments, most failures were due to computer-related issues beyond the direct contributions of the project (e.g. USB drivers, power cables, network problems etc.). Of the seven runs in 2015, one run was ended due to user intervention (a decorator powered off the robot), two due to bugs in our software, and the remaining four due to faults in software or hardware beyond our components.

The variations across deployments in terms of number of tasks completed and distance travelled were largely down to the different types of tasks performed by the robots, and the different environments they were deployed in.
For example, information serving tasks may take tens of minutes with very little travel, but door checking tasks will be brief and will also require the robot to travel both before and during the task.

\begin{figure}
\begin{center}
\includegraphics[width=\columnwidth]{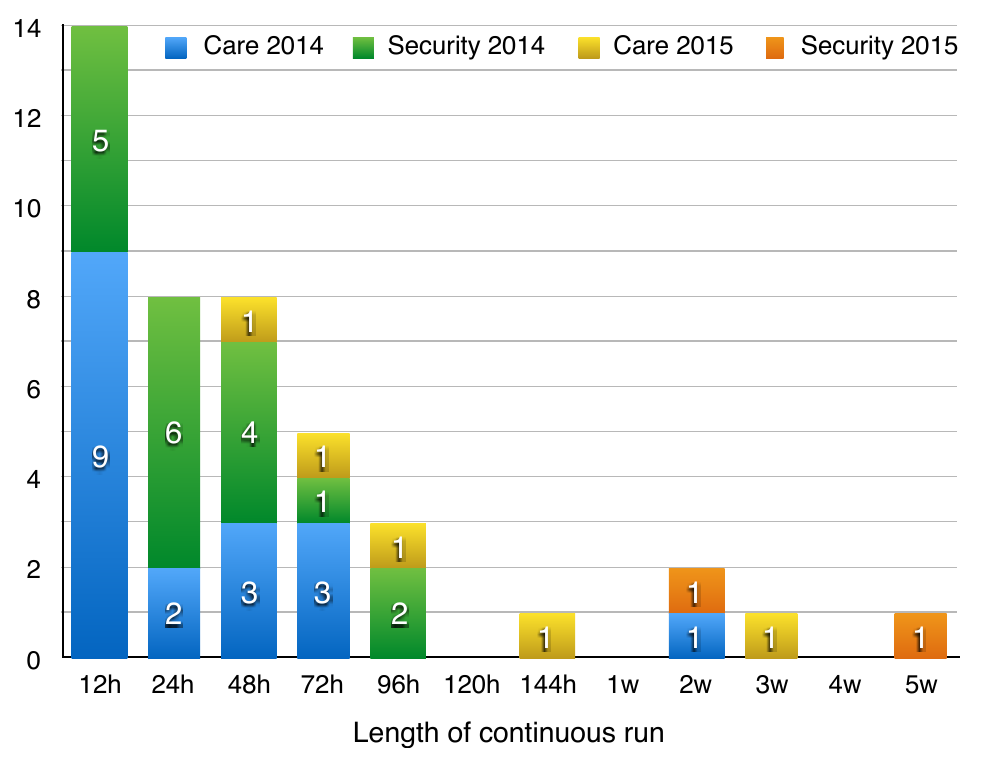}
\end{center}
\caption{A histogram of individual continuous run lengths over the 4 STRANDS deployments.\label{fig:runs}}
\end{figure}

Systems in the literature have delivered more autonomous time and distance \emph{cumulatively} (i.e. accumulated across multiple robots and/or system runs), but we believe the 28 day run is the longest a single continuous autonomous run of an indoor mobile service robot capable of multiple tasks. The most relevant comparison we can make is to the CoBots. The CoBot analysis in~\cite{1000kchallenge} reports a total of $1,279.5$ hours of autonomy time, traversing $1,006.1$km. This was achieved by four robots in $3,199$ separate continuous autonomous runs over three years, at an average of $0.31$km, $23$ minutes per run. They do not report the longest single continuous run (either in time or distance), but even an extremely long run for a CoBot would only be measured in hours, not days (as they don't have autonomous charging capabilities). In contrast the STRANDS systems performed a total of $43$ separate continuous runs, yielding a total of $2,545$ hours and $116$ km over the four deployments, at an average of $2.7$km and 58:12 hours per run. How the duration of individual runs varied can be seen in Figure~\ref{fig:runs}. Note that we use this data to provide a point of comparison. The two projects are targeting different metrics (total distance for CoBots, single run duration for STRANDS) thus the systems naturally have different performance characteristics.

Sections \ref{sec:monitored_nav} and \ref{sec:adaptive_nav} describe novel elements of our system that have enabled such long run times everyday environments. These are followed by examples of tasks that exploit these long run times to improve robot service performance.

\section{Monitored Navigation}\label{sec:monitored_nav}





Given the huge variety of situations an LTA service robot will encounter, it is impossible to develop a navigation algorithm to successfully deal with all of them.
We therefore developed a framework that executes topological navigation actions and monitors them for failure.
If a failure is detected, then the framework iterates through a list of recovery behaviours until either the navigation action completes successfully, or the list is exhausted (in which case failure is reported back to the calling component).
Failure types can be mapped to specific lists of recoveries.
When the robot's bumper is pressed, a hardware cut-off prevents it driving, therefore in this case the robot must ask to be pushed away from obstructions by nearby humans.
If the local DWA planner fails to find a path, then simply clearing the navigation costmap (to remove transient obstacles) may suffice. We also developed a backtrack behaviour which uses the PTU-mounted depth camera to sense backwards whilst reversing along the path it took to the failure location. This is triggered when navigation fails, but clearing the costmap does not overcome the failure.

\begin{table}[ht]
\centering
\begin{tabular}{p{5cm}|p{5cm}|l|l|l}
Failure				& Recoveries   &  Successful  &  Unsuccessful &  Total \\
\hline
Bumper pressed		    &  Request help via screen and voice. Repeated until recovered. &	177	& 148	& 325 \\
Navigation failure (no valid local or global path)		&  Sleep then retry; backtrack to last good pose;  &		&  &	 \\
		&  request help via screen and voice. Repeated request until recovered. &	707	& 993 & 	1700 \\
Stuck on carpet		    &  Increased velocities commanded to motors &	16 &	247	& 263 \\

\end{tabular}
\caption{Classes of navigation failure, their associated recoveries, and the overall counts of successful and unsuccessful recoveries from these failures. Per-recovery counts are show in Figure~\ref{fig:recoveries}}
\label{tab:recoveries}
\end{table}

\begin{figure*}
\begin{center}
\includegraphics[width=\textwidth]{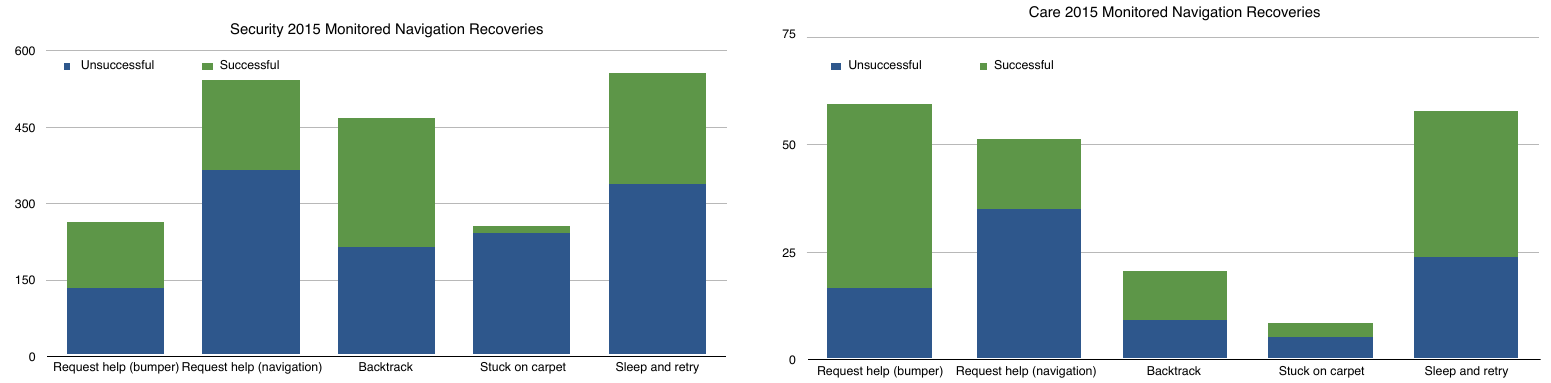}
\end{center}
\caption{Per-recovery counts for our 2015 security (left) and care (right) deployments.\label{fig:recoveries}}
\end{figure*}

\begin{figure*}
\begin{center}
\includegraphics[width=\textwidth]{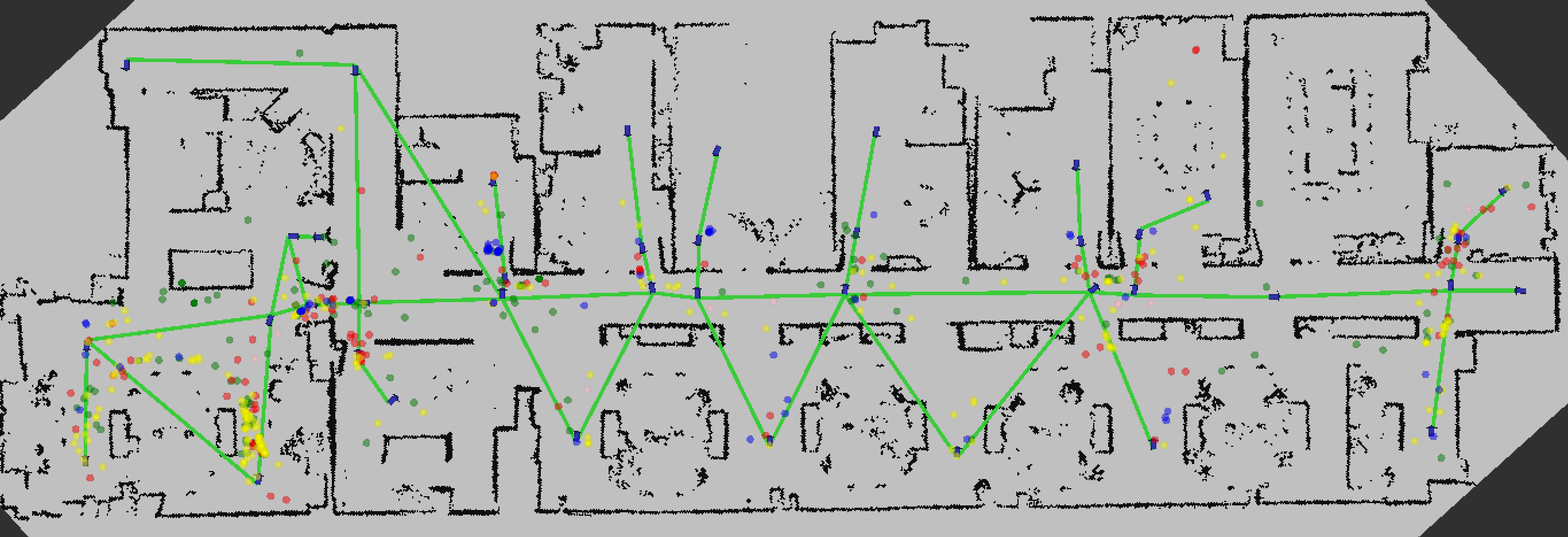}
\end{center}
\caption{The map of the deployment area in Challenge House, Tewkesbury with the topological map superimposed. Also displayed are the locations where the robot successfully recovered from a navigation failure. 
Locations where the bumper was triggered are red. The robot asked humans for help at these locations. It also did this at locations marked with green (for non-bumper fails). Places where recoveries were performed by reversing along the previous path are marked in yellow, and by simply retrying in blue.\label{fig:recovery_plot}}
\end{figure*}

Table~\ref{tab:recoveries} presents the recovery behaviours used in our 2015 deployments.
Successful recoveries are those which are not followed by another failure within one minute or one metre of travel, otherwise they are unsuccessful. 
A successful recovery may be preceded by any number of unsuccessful recoveries.
A sequence of unsuccessful recoveries can come from the monitored navigation system as it attempts recoveries that then fail, or from the task execution framework unsuccessfully trying to navigate the robot to another task after a previous failure.
Figure~\ref{fig:recovery_plot} shows where all the successful recoveries from our 2015 security deployment occurred.
They are largely clustered around areas where it was difficult to navigate, such as near doors, and close to desks. This novel approach significantly contributed to the LTA performance of our systems, as each recovered failure could have potentially caused the end of a continuous run.

\section{Adaptive Topological Navigation}\label{sec:adaptive_nav}

Whilst monitored navigation helps the robot recover from navigation failures, it does not help it avoid them. To do this we aggregate the robot's navigation experience into a Markov Decision Process (MDP) automatically built from the topological map~\cite{lacerda_iros14}. Using an MDP allows the system to model uncertainty over the success of the robot traversing an edge in the map and its expected duration. By learning models for these success probabilities and durations online, the robot is able to continually adapt its behaviour to the environment it is deployed in. Every time the robot navigates an edge, the duration and success of the traversal is logged to the robot's database. These logs are processed by FreMEn (see Section~\ref{sec:core}) to produce a temporal predictive model that allows the actions of the MDP to be assigned probabilities and travel durations appropriate for the time of execution~\cite{Mudrova:2015}. This MDP is then solved for a target location to produce a policy for topological navigation that prefers low duration edges with high success probabilities (see~\cite{lacerda_iros14} for details). This improves the system's robustness by making it avoid areas where it previously encountered navigation failures. This is only possible in an LTA setting where the robot runs repeatedly in the same environment.

\section{Predicting Human-Robot Interaction}\label{sec:inforterminal}



In the HdB care facility our robot acts as an information terminal, using its touch screen to present the date, daily menu, news etc., to staff and to residents with potentially severe dementia. This behaviour is scheduled as a task at different topological nodes in the care home. As we did not know in advance the locations and times people would prefer to interact with the robot, we allowed it to adapt its  routine based on long-term experience.
To achieve this, each node in the topological map is associated with a FreMEn model that represents the probability of someone interacting with the robot's screen at a given time. This is built from logs of screen interactions stored in MongoDB. These FreMEn models are used to predict the likelihood of interactions at given times and locations. These predictions are used by the robot to schedule where and when it should provide information during the day.

The schedule has to satisfy two contradicting objectives common to many online, active learning tasks: exploration (to create and maintain the spatio-temporal models), and exploitation (using the model to maximise the chance of interacting with people).
Exploration requires the robot to visit locations at times when the chance of obtaining an interaction is uncertain. Exploitation requires scheduling visits to maximise the chance of obtaining interactions. To tackle this trade off, the schedule is generated using Monte Carlo sampling from the location/time pairs according their FreMEn-predicted interaction probability (exploitation) and entropy (exploration).
For more details see~\cite{lifelong_exploration_ral}. 

\begin{figure}
\begin{center}
\includegraphics[width=\columnwidth]{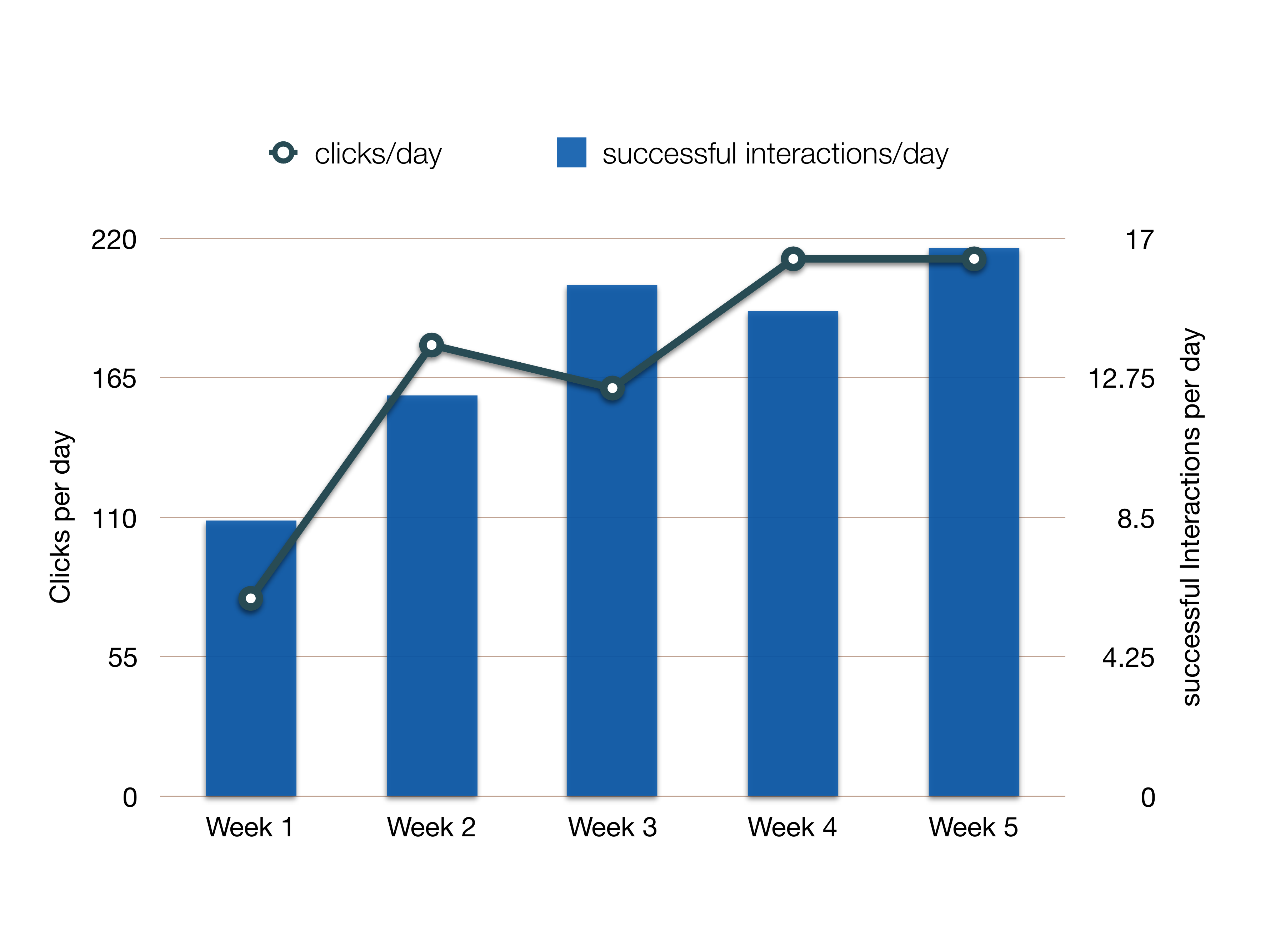}
\end{center}
\caption{The results of the robot selecting interaction times and locations using FreMEn models learnt during the 2015 care deployment.\label{fig:it-results}}
\end{figure}

Figure~\ref{fig:it-results} shows that by using this approach the robot was able to increase the number of successful interactions (i.e. when information was offered and someone interacted with the screen) on average per day over the course of its deployment. Although we have no control group to compare against, our on-site observations indicate that the robot's choices are having a positive effect. This demonstrates the ability of the system to improve its application-specific behaviour from long-term experience. 
%



\section{Activity Learning}\label{sec:activities}

\begin{figure*}
\begin{center}
\includegraphics[width=\textwidth]{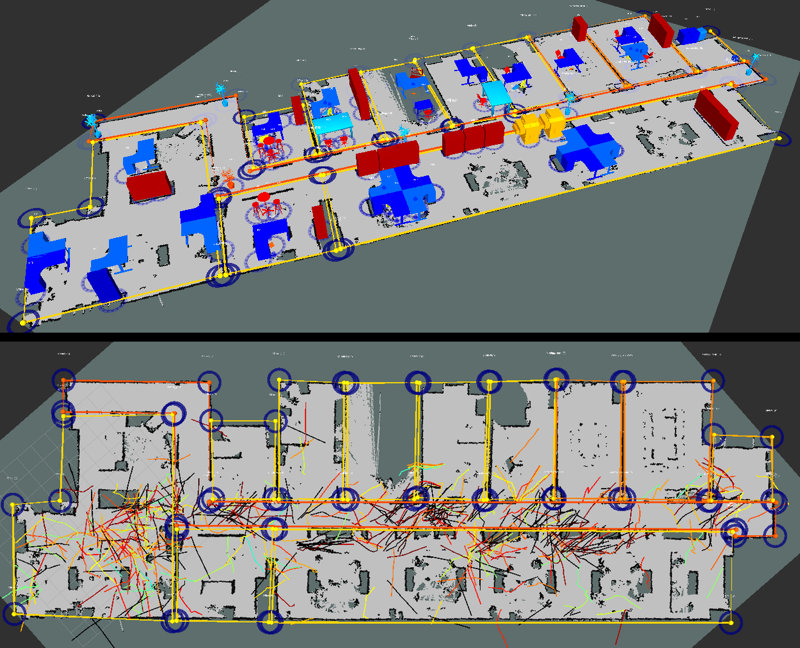}
\end{center}
\caption{Top: The manually-created semantic map from the 2015 security deployment. Bottom: Example human trajectories with length close to the average trajectory length of $2.44m$. Also pictured are the manually annotated room regions we used for task planning.\label{fig:trajectories}}
\end{figure*}

\begin{figure*}
  \begin{center}
    \includegraphics[width=0.28\linewidth]{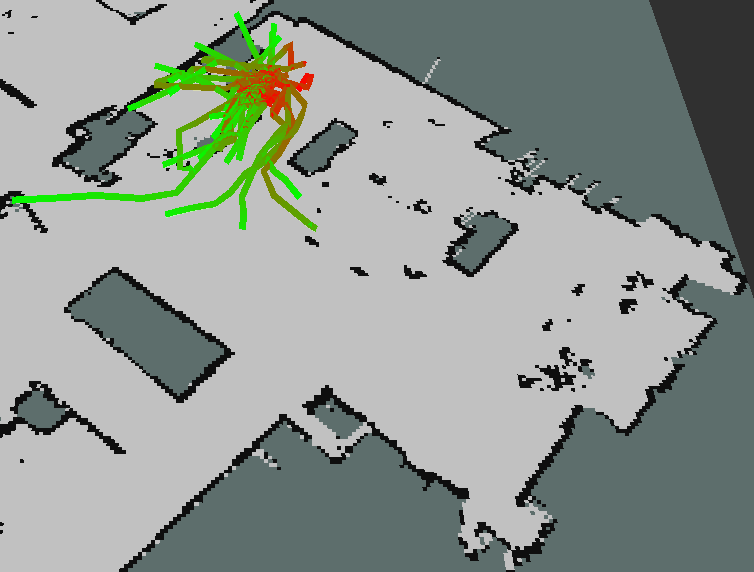} 
    \includegraphics[width=0.28\linewidth]{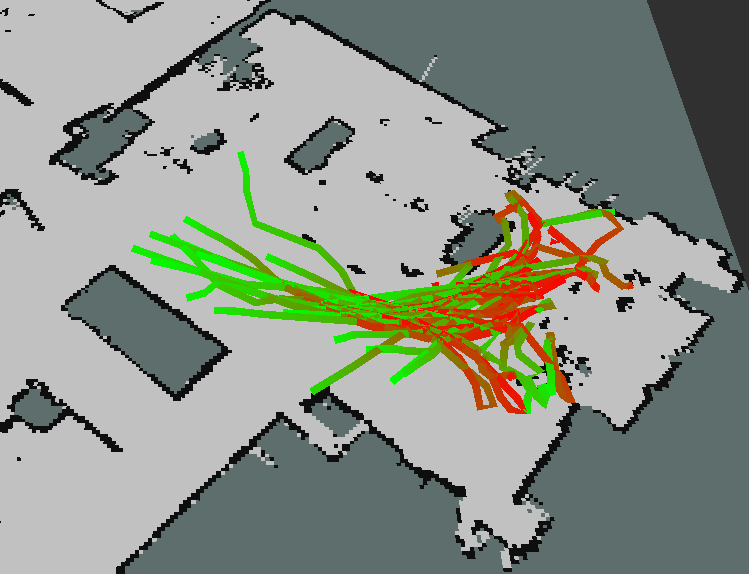} 
    \includegraphics[width=0.28\linewidth]{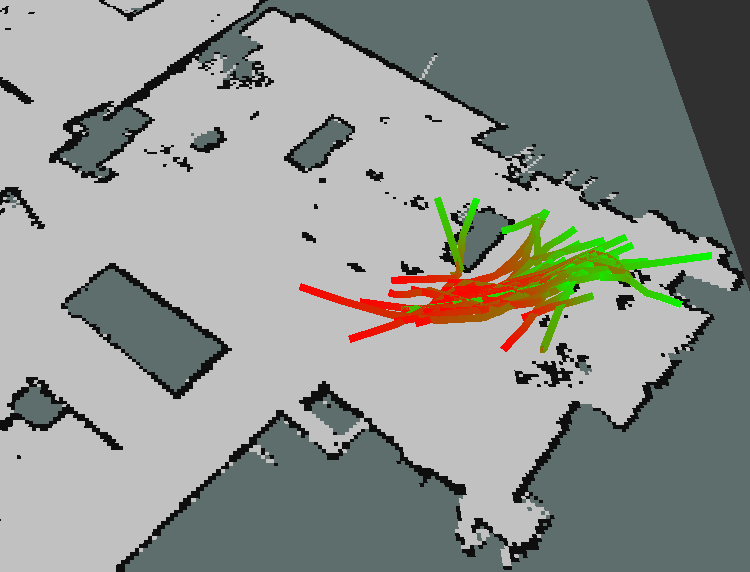} 
  \end{center}
  \caption{Trajectories belonging to three learned clusters in the region at the bottom left of Figure~\ref{fig:trajectories} (direction of motion is red to green). These can be interpreted as two clusters of a desk approaching activity, and one of desk leaving.}
  \label{fig:clusters}
\end{figure*}

In our security scenario, the robot should learn models of normal human activity then raise an alert if an observation deviates from this.
We have explored activity learning using walking trajectories (see Figure~\ref{fig:trajectories}).
Over the 2015 security deployment, the robot detected $42,850$ individual trajectories.
%
%
As described in~\cite{Duckworth:2016}, we use Qualitative Spatio-Temporal Activity Graphs (QSTAGs) to generalise from individual trajectories to spatial and temporal relations between trajectories and landmarks in a semantic map (see Figure~\ref{fig:trajectories}). A QSTAG ignores minor quantitative variations across trajectories, but captures larger, qualitative changes.
Every night the robot created QSTAGs for a subset of all trajectories (based on their displacement ratio) observed during the day. It then clustered these to create classes of movement activities. Some examples of the results can be seen in Figure~\ref{fig:clusters}.

During the day, an observation of a trajectory sufficiently far from any cluster centre triggered a task to approach the tracked human and request confirmation of their identity using a card reader. To enable a fast response it is important that the robot can accurately match the start of the trajectory to a cluster. Table~\ref{tab:traj_results} shows how the accuracy of predicting the cluster of a trajectory from an initial segment (20\%) improves as more data is gathered over the robot's lifetime. This provides another example of how a robot can improve its application-specific performance once it can operate over long periods.

\begin{table}[ht]
\centering
	\begin{tabular}{l|rrrr}
	Training weeks $(\#traj.)$ & $K$    & recall  &  prec   &  $F1$ \\
		\hline
		  week 0 (342)     &   9  & 0.24        &  0.72   &  0.29 \\
		  weeks 0-1 (511)  &  12  & 0.43        &  0.54   &  0.44 \\
		  weeks 0-2 (707)  &  12  & 0.43        &  0.56   &  0.43 \\
		  weeks 0-3 (811)  &  10  & 0.43        &  0.71   &  0.49 \\
		  weeks 0-4 (1016) &  14  & 0.48        &  0.63   &  0.53 \\
 	\end{tabular}
    \caption{Accuracy of activity cluster prediction on week 5 data, from partial input trajectories.}
	\label{tab:traj_results}
\end{table}


\section{Conclusions and Future Work}

The STRANDS Core System features a mix of design- and run-time approaches which allow it to deliver LTA in everyday environments. A key strategy for delivering long-term robustness is the monitoring of system behaviour, from the individual component level up to navigation and task behaviours, plus the ability to restart system elements on demand. This allows the system to cope with unexpected situations both internally and in the external environment. Our aim is also to use the long-term experience of failures to learn to avoid these failures in the future. We presented our approach for doing this for navigation (Section~\ref{sec:adaptive_nav}), and hope to generalise this to other parts of the system. Whilst these features provide a fundamental ability to operate autonomously for long durations in everyday environments, our robots currently have no way to manage failures which are more catastrophic, harder to predict, or both. For example, our systems have suffered from PC component failure and subtle networking issues. In the future we would like to look at the use of redundancy and online reconfiguration (e.g. substituting a failing software or hardware component), coupled with more general failure detection approaches (both have which have been extensive researched in robots and other systems). 

Our robots are able to learn online from lengths of experiences that no other robots to date have access to. The results above demonstrate what we have always known from machine learning: more data improves performance. However, the novel element here is that a robot must be able to operate for longer in order to gather additional data, and can make active choices about what data is gathered. 

%
%
%
%
%
%
In the future we will also focus on the robot's ability to understand human activities (the major causes of environment dynamics at most scales) and to actively close the gaps in its understanding it has already obtained from weeks of autonomous runtime.

\section{Acknowledgements} 

We would like to acknowledge the contribution our project reviewers and project officers have made to our research: Luc De Raedt, James Ferryman, Horst-Michael Gross, Olivier Da Costa and Juha Heikkil\"a. The research leading to these results has received funding from the European Union Seventh Framework Programme (FP7/2007-2013) under grant agreement No 600623, STRANDS.

\bibliographystyle{IEEEtran}
\bibliography{lta}
\end{document}